\documentclass[lettersize,journal]{IEEEtran}
\usepackage{amsmath,amsfonts}
\usepackage{algorithmic}
\usepackage{algorithm}
\usepackage{array}
\usepackage[caption=false,font=normalsize,labelfont=sf,textfont=sf]{subfig}
\usepackage{textcomp}
\usepackage{stfloats}
\usepackage{float}
\usepackage{url}
\usepackage{verbatim}
\usepackage{graphicx}
\usepackage[table]{xcolor}
\usepackage{booktabs}
\usepackage[colorlinks, linkcolor=blue, anchorcolor=blue, citecolor=blue]{hyperref}
\usepackage{cite}
\usepackage{multirow}
\usepackage{adjustbox}
\usepackage{graphicx}

\begin{document}

\title{PolUR: Polarization Aware Multimodal Degradation Perception for Underwater Image Restoration}

\author{Wenshuo Zhang, Xiangyue Zhang*, Junlin Li, Wenjie Tian, Yubo Zheng, Hongli Xu, Jingyu Ru, Xinde Li**, Senior Member, IEEE 
\thanks{This work was supported in part by the Natural Science Foundation of China under Grant U25A20477, Ministry of Education Discipline Project (No. JYB2025XDXM209)\par
\textit{(Corresponding authors: Xiangyue Zhang; Xinde Li.)}}
\thanks{Wenshuo Zhang, Xiangyue Zhang Junlin Li, Yubo Zheng, Wenjie Tian, Yubo Zheng, Hongli Xu, Jingyu Ru and Xinde Li are with the Faculty of Robot Science and Engineering, Northeastern University, Shenyang 110819, China (email: zhangxiangyue@mail.neu.edu.cn, lixinde@mail.neu.edu.cn).}}




\maketitle

\begin{abstract}
Recovering high-quality images from degraded underwater observations is crucial for underwater exploration.
Polarization observations show promise as an effective source of complementary physical cues for recovering target radiance under clear-water conditions.
However, existing methods have yet to jointly model raw intensity information and polarization cues to enable image restoration driven by an understanding of mixed degradations (color casts, haze, and low light).
To overcome these limitations, we propose PolUR, a unified multimodal language-driven framework that jointly uses polarization and RGB information to identify degradation types and severity levels through structured semantic instructions and enhances joint restoration representations under the guidance of degradation-aware semantic features to improve restoration robustness under mixed degradations.
We further construct a new benchmark named MixPolar, which contains underwater images exhibiting real-world degradations and their corresponding degradation descriptions. MixPolar supports systematic evaluation of degradation perception, image restoration, and performance on downstream tasks.
Extensive experiments demonstrate that PolUR effectively integrates polarization and RGB information, achieving competitive performance in degradation perception and image restoration, while also demonstrating generalization capability and utility for downstream tasks.
\end{abstract}


\section{Introduction}
\begin{figure}[!t]
\centering
\includegraphics[width=3.3in]{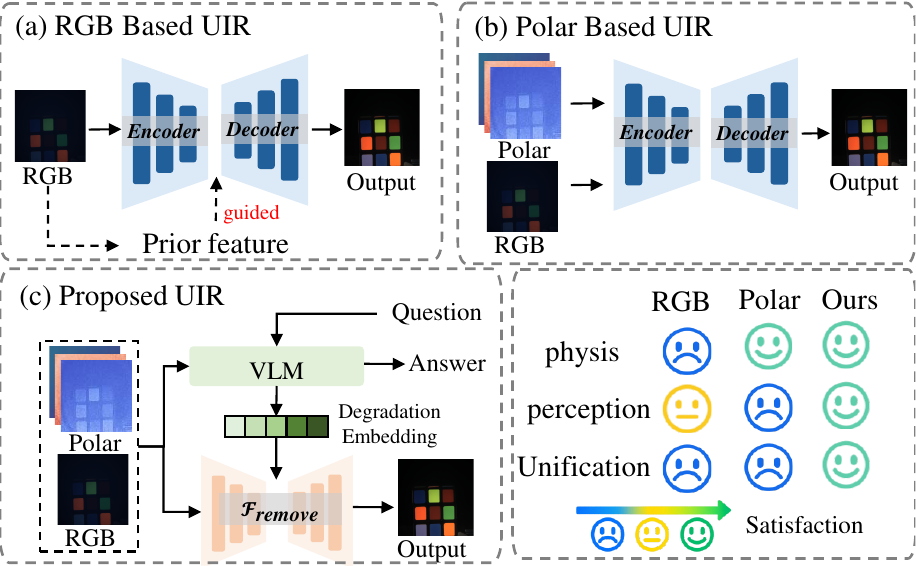}
\caption{Comparison between proposed PolUR and previous approaches to underwater image restoration. (a) RGB-based UIR methods rely primarily on degraded intensity observations, with degradation priors inferred from the same information-limited RGB inputs, which restricts reliable degradation perception under complex underwater conditions. (b) Polarization-based UIR methods exploit physical cues related to light propagation and scattering, but incorporating these cues alone does not ensure adequate characterization of water conditions and compound degradation states. (c) PolUR unifies physical observation, degradation perception, and restoration decision-making within a framework, jointly using RGB and polarization cues to reason about coexisting degradations and guide restoration.}

\label{motivation}
\end{figure}
\IEEEPARstart{U}{nderwater} imaging is significantly affected by light absorption and scattering, which attenuate target radiance and introduce scattered-light interference, resulting in color casts, low contrast, and poor visibility~\cite{akkaynak2017space}. 
Optical differences between water and air limit the direct transfer of restoration methods to underwater environments~\cite{li2022all, R2R, fan2026degradation}, motivating dedicated underwater image restoration (UIR) techniques that enhance visual quality and make scene information more discernible, thereby providing more reliable observations for downstream vision tasks.

Traditional UIR methods rely on statistical priors~\cite{traditional_2,traditional_4} or physical models~\cite{traditional_1,traditional_3}. However, their reliance on predefined assumptions and sensitivity to parameter estimation accuracy can limit their generalization to complex underwater scenes.

Existing RGB-based methods for underwater image restoration typically enhance feature representations through multi-domain modeling~\cite{SSUIE,wang2025mcsf} and prior guidance~\cite{cao2025unveiling,yao2026conditional,fan2025llava,SLURPP,DIE2UIE,uniuir}.
Multi-domain approaches combine domain transformations with cross-domain feature fusion, with their adaptability to degradations relying primarily on architecture-driven adaptive feature learning.
To enable more targeted control over restoration, recent studies increasingly incorporate priors from generative models and vision--language models for condition-guided feature modulation (as shown in Fig.~\ref{motivation}(a)).
However, the loss of scene information caused by absorption and scattering in water is difficult to fully compensate for through feature modeling and prior-based inference alone, leaving their performance constrained by the original observations.\par
In contrast, polarization imaging captures polarization variations induced during underwater light propagation and scattering, thereby characterizing the differences between scattered light and target radiance and providing a physical basis for separating and suppressing scattering components~\cite{polarrecovery}.
Meanwhile, the polarization responses of target reflections vary with the material and surface characteristics of underwater objects, revealing additional intrinsic information about the targets. Such characteristics provide a more informative basis for underwater image restoration.\par
Existing polarization-based UIR methods mainly rely on scattering models, using multi-angle polarization observations to estimate backscattering and transmission map for physically constrained radiance recovery~\cite{polarrecovery,treibitz2008active,schechner2004clear}. However, their reliance on idealized imaging assumptions limits robustness and generalization under complex real-world underwater conditions. With advances in deep learning, data-driven restoration methods that exploit the physical relationship between target radiance and polarization information have attracted increasing attention. 
Specifically, as shown in~\ref{motivation}(b), these methods jointly use RGB and polarization information to learn a mapping from degraded observations to target radiance through improvements in network architectures or restoration pipelines~\cite{hu2020polarimetric,UPI2Diff,UPGD}. 
However, imaging characteristics vary across aquatic environments, as do the relative contributions of the two modalities. 
Focusing solely on incorporating polarization information without adequately representing water conditions limits the model’s ability to adapt to varying scene conditions.

To address these challenges, we propose PolUR, an instruction-driven framework for polarization-based underwater image restoration that unifies physical perception and language understanding, as illustrated in Fig.~\ref{motivation}(c). The framework takes RGB and polarization information together with textual instructions as input and performs degradation diagnosis and instruction-guided image restoration.

The framework comprises two components: a Polarization-aware Degradation Understanding Model (PolarDU) and a Unified Response-Guided Restoration Network (UR$^2$Net). PolarDU combines the complementary degradation representations of RGB and polarization observations with structured text to enable explicit perception and understanding of complex degradations. Guided by degradation-aware hidden states extracted from PolarDU, UR$^2$Net adapts the restoration process to different degradation conditions. We further introduce an RGB--polarization alignment (RPA) module and a progressive training strategy to narrow the representation gap between the two modalities and facilitate cross-modal feature fusion.
To support systematic joint degradation analysis and restoration evaluation, we construct MixPolar, an RGB--polarization benchmark covering diverse underwater degradation types and severity levels.

The main contributions of this work are summarized as follows:

1) We introduce MixPolar, an RGB--polarization benchmark covering diverse types and severity levels of mixed underwater degradations, with a unified evaluation pipeline spanning degradation analysis, image restoration, and downstream vision tasks to assess both restoration quality and the utility of restored images.

2) We propose PolUR, the first unified framework that combines RGB and polarization observations with structured textual instructions for joint degradation understanding and uses the resulting degradation-aware representations to guide restoration under mixed underwater degradations.

3) We propose a progressive training strategy and RPA module to gradually align semantic and physical representations while preserving the model’s visual understanding and image restoration capabilities.

\section{Related Work}
This section reviews RGB-based and polarization-based underwater image restoration, as well as all-in-one image restoration in both general and underwater scenes.

\subsection{RGB-based Underwater Image Restoration}
RGB-based underwater image restoration methods primarily learn enhanced representations from degraded observations to establish a stable restoration mapping~\cite{RGB1}\cite{RGB2}.

Some methods construct enhanced representations through multi-domain feature modeling based on degraded observations. Peng et al.~\cite{SSUIE} jointly model degradation features in the spatial and frequency domains, while Wang et al. fuse complementary features from the RGB and XYZ color spaces to improve color restoration and detail enhancement.

Other methods derive vision or language priors from degraded RGB observations to guide representation enhancement. 
Wu et al.~\cite{SLURPP} leverage generative models to jointly restore the clear scene and estimate transmission and backscattering from a single RGB image. 
Cao et al.~\cite{cao2025unveiling} use high-level semantic representations from CLIP to construct perceptual supervision constraints for training. 
Yao et al.~\cite{yao2026conditional} identify degradation categories in RGB inputs through CLIP and use the corresponding representations to guide restoration. 
Wang et al.~\cite{DIE2UIE} combine CLIP-derived degradation representations with RAM-generated semantic tags to guide adaptive restoration. 
Fan et al.~\cite{fan2025llava} use LLaVA to extract the textual descriptions from input images to refine diffusion conditions.

Despite improvements in restoration performance, these methods still rely primarily on degraded RGB observations for scene-specific restoration, leaving their reliability constrained by information loss caused by absorption and scattering.

\subsection{Polarization-based Underwater Image Restoration}
Polarization-based underwater image restoration leverages polarization cues associated with underwater light propagation and scattering to facilitate scene radiance recovery. Existing approaches can be broadly divided into physics-driven and learning-based methods, with the latter further characterized by advances in network architecture and restoration paradigm.

The Jaffe–McGlamery model~\cite{JMmodel1,JMmodel2} established the physical basis for modeling underwater image formation.
Schechner et al.~\cite{polarrecovery}\cite{treibitz2008active} exploited multi-angle polarization observations to characterize the polarization differences between target radiance and scattered light, thereby enabling scattering suppression under general polarization assumptions. Li et al.~\cite{li2024underwater} further developed a polarization imaging optimization model and employed local parameter optimization to improve the stability of parameter estimation.

With the development of deep learning, data-driven polarization representation learning has gradually become an important direction in underwater polarization restoration.

At the architectural level, Hu et al.~\cite{hu2020polarimetric} employed a densely connected network to fuse multi-level polarization features and learn the nonlinear mapping between polarization observations in turbid water and clear-scene radiance. Lin et al.~\cite{lin2025transformer} introduced CNN and Transformer modules into U-Net to model local details and long-range dependencies, respectively, thereby enhancing target reconstruction under turbid conditions. Fan et al.~\cite{fan2025meta} further combined deformable convolution with meta-learning to improve adaptation under limited-data settings.

At the restoration paradigm level, Shen et al.~\cite{u2pnet} adopted a dual-branch network to estimate the transmission map and background light, respectively, and incorporated no-reference constraints to enable unsupervised polarization-based image restoration. Liu et al.~\cite{UPI2Diff} introduced diffusion models into underwater polarization restoration and employed a dual-stream generation process to separately model image intensity and polarization information, enabling their joint restoration under turbid conditions. Ju et al.~\cite{UPGD} designed a lightweight polarization feature extraction module and injected polarization information as auxiliary features into existing RGB-based underwater enhancement networks, thereby extending the integration of polarization cues into general restoration frameworks.

These studies have advanced the use of polarization information in underwater image restoration. Building on these advances, further research is warranted on representing compound degradation states in complex underwater environments and using these representations to guide RGB–polarization fusion and image restoration.

\subsection{All-in-One Image Restoration}
To address unknown degradation types, dynamically varying degradation severity, and the coexistence of multiple degradations in real-world scenarios, All-in-One image restoration aims to adaptively handle diverse degradations within a unified framework. 
Li et al.~\cite{li2022all} employed contrastive learning to construct a degradation encoder and used degradation-guided restoration to accommodate unknown degradations of varying types and severity levels.
Wang et al.~\cite{R2R} constructed a retrieval-based degradation bank to dynamically retrieve degradation priors matched to the input, allowing external degradation knowledge to participate in the restoration process. 
Fan et al.~\cite{fan2026degradation} leveraged a pretrained vision-language model to capture high-level degradation semantics and combined degradation-aware prompting with feature compensation to adaptively guide restoration under diverse degradation conditions.

However, the distinct degradation characteristics caused by underwater absorption and scattering limit the direct transfer of general all-in-one restoration methods to underwater scenes.

In underwater scenes, Zhang et al.~\cite{uniuir} employed a mixture-of-experts mechanism to model different degradation patterns, enabling adaptive restoration of heterogeneous and spatially varying underwater degradations.
Nevertheless, its performance remains constrained by limited coverage of degradation conditions. Moreover, its depth priors are inferred solely from degraded RGB images, the pretrained depth estimator faces a domain gap when applied to underwater scenes.

\section{Proposed Method}
In the following sections, we first introduce the construction of polarization representations (Section~\ref{subsec:3A}) and provide an overview of the proposed framework (Section~\ref{subsec:3B}). 
We then elaborate on the two components of our method, including UR$^2$Net (Section~\ref{subsec:3C}) and PolarDU (Section~\ref{subsec:3D}). Finally, we present the training Strategy (Section~\ref{subsec:3E}).

\subsection{Polarization Representation Construction}
\label{subsec:3A}
Given the observations captured by a division-of-focal-plane (DoFP) polarization camera at four analyzer orientations $0^{\circ}$, $45^{\circ}$, $90^{\circ}$, $135^{\circ}$, denoted by $I_{0}$, $I_{45}$, $I_{90}$, $I_{135}$, the linear Stokes components are computed as~\cite{mei2022glass,tang2024learning}:
\begin{equation}
\begin{bmatrix}
S_{0}\\
S_{1}\\
S_{2}
\end{bmatrix}
=
\begin{bmatrix}
\dfrac{I_{0}+I_{45}+I_{90}+I_{135}}{2}\\[6pt]
I_{0}-I_{90}\\[4pt]
I_{45}-I_{135}
\end{bmatrix}.
\label{eq:stokes_components}
\end{equation}
where $S_{0}$ represents the total intensity, while $S_{1}$ and $S_{2}$ characterize the polarization differences between the horizontal--vertical and diagonal analyzer pairs, respectively.\par
\begin{figure}[htbp!]
\centering
\includegraphics[width=3.3in]{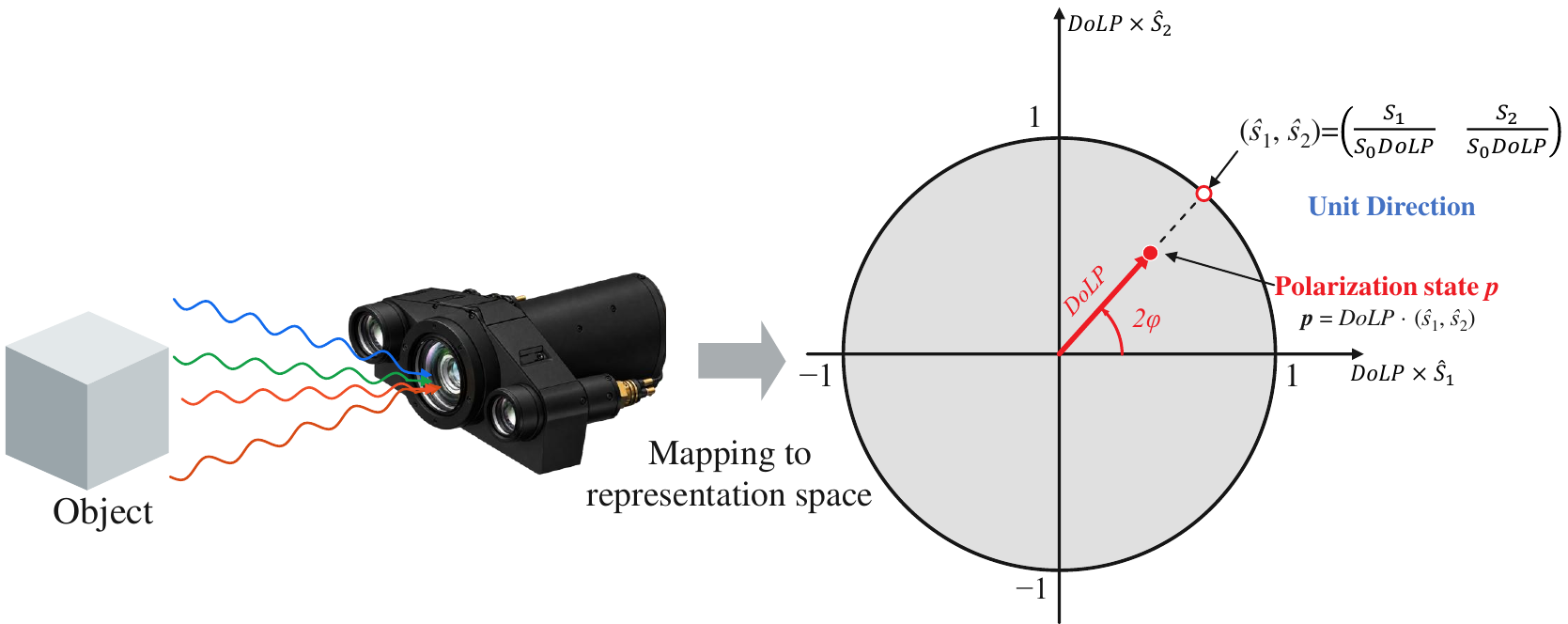}
\caption{Polarization Observation-to-Representation Mapping.}
\label{Mapping}
\end{figure}
As shown in Fig.~\ref{Mapping}, we represent the normalized linear polarization state by jointly encoding the polarization magnitude with $\operatorname{DoLP}$ and the polarization orientation with the unit-circle coordinates $(\hat{S}_{1},\hat{S}_{2})$. The resulting polarization representation $\mathbf{P}=[\mathrm{DoLP},\hat{S}_{1},\hat{S}_{2}]$ can be directly derived from the Stokes components as
\begin{equation}
\mathbf{P} =
\left[
\frac{\sqrt{S_1^2+S_2^2}}{S_0},
\mathcal{M}_{\varepsilon}\!\left[
\frac{S_1}{S_0\operatorname{DoLP}}
\right],
\mathcal{M}_{\varepsilon}\!\left[
\frac{S_2}{S_0\operatorname{DoLP}}
\right]
\right].
\label{eq:polar_rep}
\end{equation}
Here, $\mathcal{M}_{\varepsilon}[\cdot]$ sets the two directional components to zero wherever $\sqrt{S_1^2+S_2^2}\le\varepsilon$ to suppress unreliable orientation cues from weak linearly polarized signals. Constructed directly from the Stokes components, this representation retains informative polarization cues for subsequent feature learning without explicit estimation of the angle of polarization (AoP).

\subsection{Overall Framework}
\label{subsec:3B}
\begin{figure*}[htbp!]
\centering
\includegraphics[width=7in]{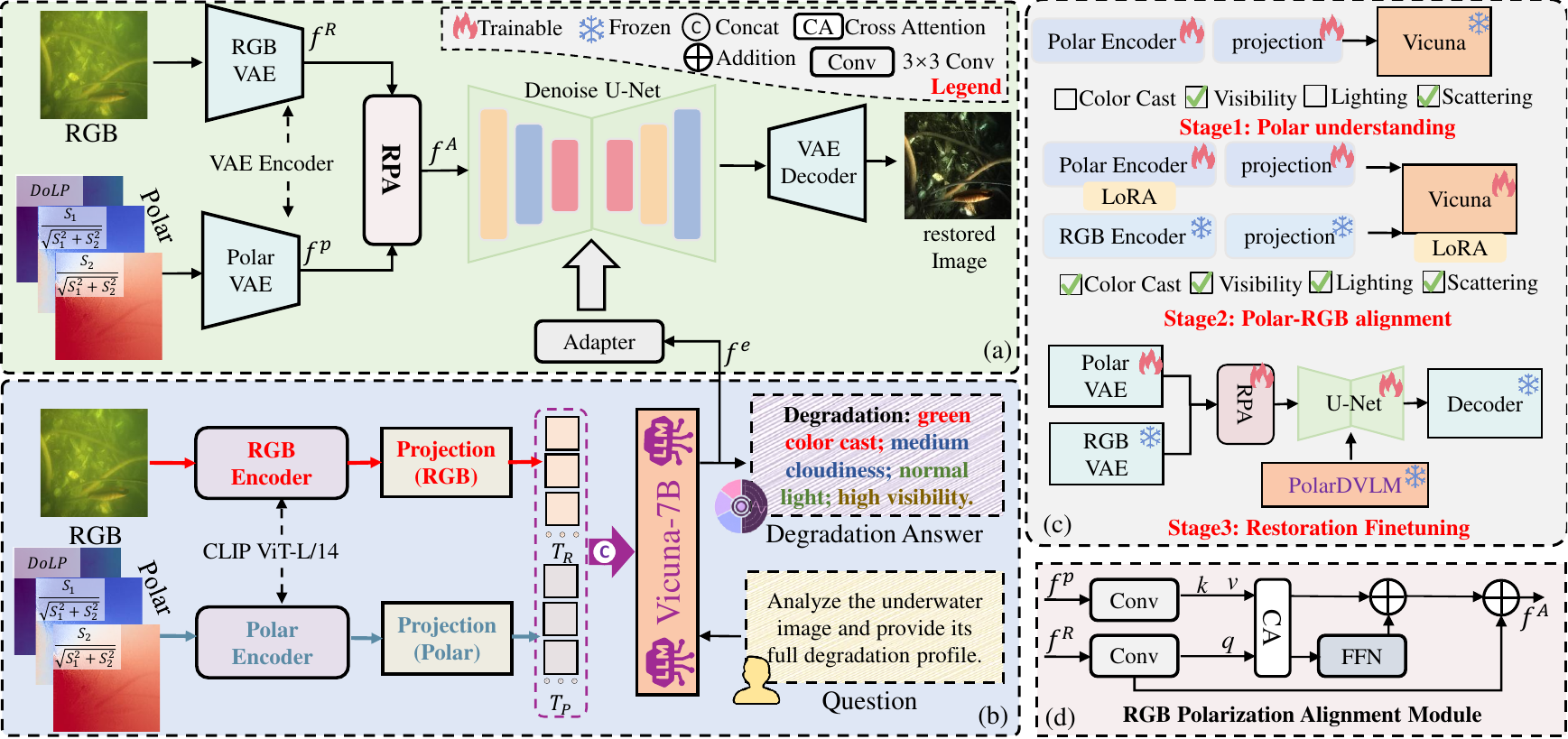}
\caption{Overview of the proposed PolUR framework. (a) and (b) jointly constitute the inference pipeline. Given an RGB image together with $\mathbf{P}$, PolUR produces both an explicit degradation response and a restored image. Specifically, (a) represents UR$^2$Net, which performs response-guided image restoration, while (b) denotes the proposed PolarDU for multimodal degradation understanding. (c) illustrates the overall training procedure of PolUR. (d) presents the RPA module embedded in UR$^2$Net, which aligns and adaptively integrates the polarization and RGB latent representations.
}
\label{Overview}
\end{figure*}

As illustrated in Fig.~\ref{Overview}, PolUR comprises two main components: (a) UR$^2$Net and (b) PolarDU.
Given the degraded RGB image $I$, the polarization representation $\mathbf{P}$, and a degradation-assessment question $Q$, PolarDU generates a degradation description $\hat{A}$ that identifies the color cast and specifies the scattering severity, illumination level, and visibility level. The contextual representations of its answer tokens, denoted by $f^e$, provide semantic guidance for restoration. 
UR$^2$Net receives $I$, $\mathbf{P}$, and the degradation-aware semantic feature $f^e$ as inputs. The inputs $I$ and $\mathbf{P}$ are encoded into latent features $f^R$ and $f^p$ by two independent VAE encoders~\cite{kingma2013auto}. To bridge the modality gap, $f^R$ and $f^p$ are mapped into a shared representation space and fused via RPA to obtain a unified enhanced feature $f^a$. The resulting feature $f^a$ is then fed into the denoising U-Net, whose operation is modulated by $f^G$ after transformation through a semantic adapter, yielding the restored image $\hat{I}$.

\subsection{Unified Response-Guided Restoration Network}
\label{subsec:3C}
RGB and polarization information differ in their ability to characterize underwater degradations, while compound degradations complicate the selection of targeted restoration strategies. To address these challenges, we propose UR$^2$Net, a single-step latent diffusion framework that adaptively fuses the latent representations of both modalities and incorporates degradation semantics encoded in the hidden states of VLM responses to guide restoration, enabling unified restoration across diverse underwater scenes without iterative reverse-diffusion sampling.

As shown in Fig.~\ref{Overview}(a), $I$ and $\mathbf{P}$ are encoded into $f^R$ and $f^p$, respectively, by two independent VAE encoders initialized with the pretrained encoder weights of the Stable Diffusion 2 VAE. These representations are then fused by the RPA module shown in Fig.~\ref{Overview}(d) as follows:
\begin{equation}
f^A = \widetilde{f}^{r}+\operatorname{CA}(\widetilde{f}^{r},\widetilde{f}^{p})+\operatorname{FFN}(\operatorname{CA}(\widetilde{f}^{r},\widetilde{f}^{p})),
\end{equation}
where $\widetilde{f}^{r}$ and $\widetilde{f}^{p}$ are obtained by projecting $f^r$ and $f^p$ through $3\times3$ convolutions. CA denotes cross-attention over $3\times3$ neighborhoods, and FFN denotes a $1\times1$ feed-forward projection.

The fused features $f^A$ are fed into a restoration network based on the U-Net architecture of Stable Diffusion 2 (SD2)~\cite{rombach2022high}. Within the U-Net, the projected semantic features $f^E$ interact with image features through cross-attention layers in the downsampling, middle, and upsampling paths, providing degradation-aware guidance for reconstruction. Together, $f^A$ and $f^E$ serve as the spatial and semantic conditions for learning a direct restoration mapping. A single-step formulation is adopted, in which the U-Net is evaluated once at a fixed terminal timestep $\tau$, conditioned on these features. Its output is then converted into the restored latent representation using the $v$-prediction parameterization with $z_\tau=0$, followed by VAE decoding to reconstruct the restored image.
\subsection{Polarization-aware Degradation Understanding Model}
\label{subsec:3D}
Directly expanding the RGB input channels to incorporate physical information may disrupt the semantic representations learned during pretraining. To introduce physical awareness while preserving the base reasoning capabilities, a degradation-aware architecture built on LLaVA-1.5~\cite{llava} is proposed, as shown in Fig.~\ref{Overview}(b). Feature extraction from intensity and physical information is decoupled, with the intensity branch providing a stable semantic reference and the physical branch supplying complementary degradation cues.
Given $I$ and $\mathbf{P}$, feature extraction and projection into the language model's embedding space are jointly expressed as
\begin{equation}
\widetilde{T}_R=\mathcal{F}_R(I),\qquad
\widetilde{T}_P=\mathcal{F}_P(\mathbf{P}),
\end{equation}
where $\mathcal{F}_R$ and $\mathcal{F}_P$ denote modality-specific encoding and projection mappings, and $\widetilde{T}_R$ and $\widetilde{T}_P$ are the RGB and polarization tokens in the embedding space.

As $\widetilde{T}_R$ primarily encodes appearance-related information while $\widetilde{T}_P$ provides complementary polarization-aware physical cues, we preserve them as separate token sequences for degradation understanding. Different from the restoration branch, where local spatial correspondences between the two modalities are explicitly modeled through cross-attention, the understanding branch concatenates $\widetilde{T}_R$ and $\widetilde{T}_P$ with the text embeddings of $Q$, allowing the LLM to selectively aggregate complementary cross-modal evidence through self-attention and autoregressively generate the degradation description $\hat{a}$.
To transfer this degradation understanding to the restoration branch, we directly use the contextualized hidden states associated with $\hat{a}$. Passing only the discrete response text to restoration would omit the internal representations formed during multimodal reasoning. These hidden states retain response semantics conditioned on the visual observations and the query, providing semantic guidance for restoration. Specifically, the final-layer hidden states corresponding to valid generated answer tokens are stacked to construct the semantic representation:
\begin{equation}
f^e=
\left[\mathbf{h}^{(L)}_t\right]_{t\in\mathcal{A}_e}
\in\mathbb{R}^{T_e\times d_\ell},
\end{equation}
where $\mathbf{h}^{(L)}_t$ denotes the final-layer hidden state at position $t$, $\mathcal{A}_e$ is the set of valid answer-token positions, and $T_e=|\mathcal{A}_e|$. Prompt, end-of-sequence, and padding positions are excluded.

\subsection{Training Strategy and Objectives}
\label{subsec:3E}
We develop a three-stage progressive training strategy, as illustrated in Fig.~\ref{Overview}(c).In the first stage, the RGB branch is bypassed, and polarization-only alignment is performed to associate polarization cues with degradation semantics. In the second stage, RGB observations are introduced for joint RGB–polarization training. LoRA adapters~\cite{hu2021lora} are employed to adapt the polarization encoder and language model for cross-modal degradation understanding while keeping their pretrained base weights frozen. Both recognition stages are trained using the cross-entropy loss:
\begin{equation}
\mathcal{L}_{\mathrm{rec}}
=
-\frac{1}{T}\sum_{t=1}^{T}
\log p_{\theta}\!\left(a_t\mid a_{<t},Q,\mathcal{V}\right),
\end{equation}
where $a_t$ is the $t$-th token of the target response $a$, and $a_{<t}$ denotes the preceding answer tokens. The target response length $T$ includes the end-of-sequence token. The visual context $\mathcal{V}$ consists of $\widetilde{T}_P$ in the first stage and both $\widetilde{T}_R$ and $\widetilde{T}_P$ in the second stag.

In the third stage, the degradation recognizer is frozen, and only $f^E$ is used to provide semantic guidance for image restoration. The RGB VAE encoder and decoder also remain frozen, while the remaining restoration components are trained. Restoration losses are computed between the restored image $\hat{I}$ and the paired reference image $I_{\mathrm{gt}}$, with subject masks used for structural supervision:
\begin{equation}
\mathcal{L}_{\mathrm{rest}}
\lambda_{1}\mathcal{L}{1}
+
\lambda{s}\mathcal{L}{\mathrm{SSIM}}^{\mathrm{subj}}
+
\lambda{p}\mathcal{L}{\mathrm{LPIPS}},
\end{equation}
where $\lambda{1}$, $\lambda_{s}$, and $\lambda_{p}$ weight the three loss terms. $\mathcal{L}_{1}$, $\mathcal{L}_{\mathrm{SSIM}}^{\mathrm{subj}}$, and $\mathcal{L}_{\mathrm{LPIPS}}$ denote the pixel-wise $\ell_1$ loss, subject-mask-weighted SSIM loss, and LPIPS perceptual loss.

\section{Experiment}
\subsection{Implementation Details}
AdamW~\cite{loshchilov2017fixing} is used as the optimizer for all three training stages. In the first stage, we set the learning rates for the polarization input layer and projector to $5\times10^{-5}$ and $2\times10^{-4}$, and train for 10 epochs with a batch size of 4 and 4 gradient accumulation steps. In the second stage, the learning rates for the LoRA adapters, Vicuna LoRA adapters, and projector are set to $2\times10^{-5}$, $2\times10^{-4}$, and $5\times10^{-5}$, respectively. Training lasts for 20 epochs with a batch size of 4 and 2 gradient accumulation steps. In the final stage, we set the learning rate to $5\times10^{-6}$ for both the polarization VAE and language adapter module, $5\times10^{-5}$ for the fusion module, and $1\times10^{-5}$ for the U-Net. This stage lasts for 15 epochs with a batch size of 1 and 4 gradient accumulation steps. All training is implemented in PyTorch and conducted on an NVIDIA A40 GPU. The input image resolution in the final stage is set to $512\times512$.
\subsection{Experiment Settings}
\begin{figure}[htbp!]
\centering
\includegraphics[width=3.3in]{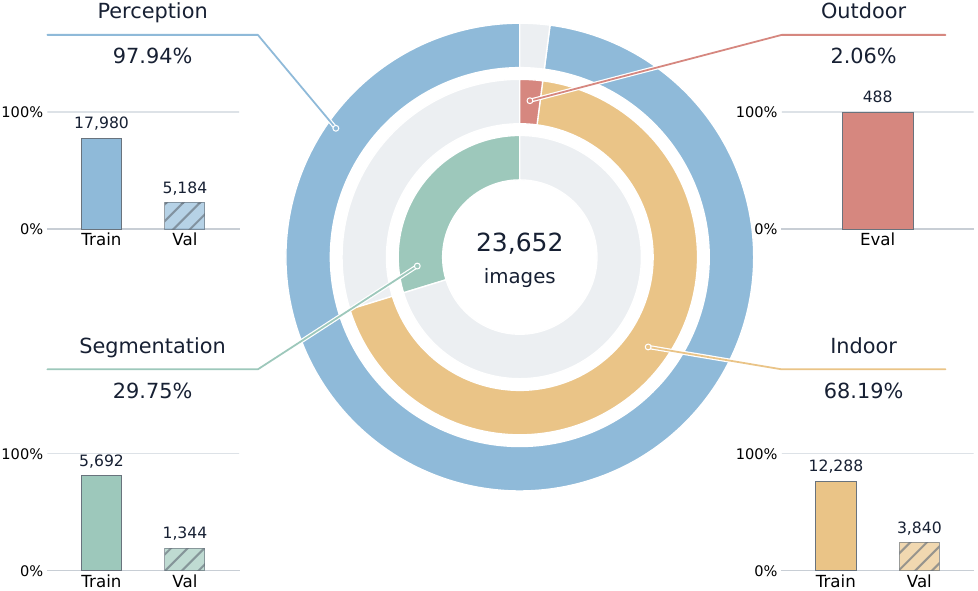}
\caption{The distribution of the Mixpolar.}
\label{distribution}
\end{figure}
\textit{1) Details on MixPolar:}
To support the training and evaluation of PolUR, we construct MixPolar, a multi-task polarization dataset, by integrating and reprocessing raw polarization measurements from existing datasets~\cite{ma2025stamf,li2025rgbpuie,zhang2026p} and providing task-specific annotations. As shown in Fig.~\ref{distribution}, MixPolar supports degradation perception, image restoration, and downstream task evaluation.
For degradation perception, we annotate four attributes: color cast, scattering severity, illumination conditions, and overall visibility, and convert the resulting labels into standardized textual descriptions. Scattering severity is categorized as low, medium, or high based on the corresponding optical thickness values. The remaining labels are assigned or verified through manual visual assessment. Specifically, color cast is categorized as none, red, green, or blue; illumination conditions as low-light or normal; and overall visibility as low, medium, or high. 
For image restoration, we establish an Indoor subset with paired ground-truth images acquired under laboratory conditions and an Outdoor subset without paired reference images.
For downstream task evaluation, we use images annotated with binary segmentation masks and compare segmentation performance on the original and restored images to assess the impact of restoration on downstream visual tasks.

\textit{2) Evaluation Metrics:} We evaluate degradation perception using mean macro-F1 and joint accuracy. Mean macro-F1 provides a balanced assessment of recognition performance across degradation attributes and classes, while joint accuracy assesses the reliability of complete degradation diagnoses. For indoor image restoration, we use PSNR, SSIM~\cite{wang2004image}, LPIPS~\cite{zhang2018unreasonable}, and DISTS~\cite{ding2020image} to assess reconstruction fidelity, structural similarity, and perceptual similarity. For outdoor evaluation, where paired references are unavailable, we use the no-reference metrics UIQM~\cite{panetta2015human}, URanker~\cite{guo2023underwater}, and UCIQE~\cite{yang2015underwater} to assess image quality.\par

\subsection{Effectiveness of Polarization Information}
\label{sec:polarization_effectiveness}
we investigate the response of $\mathbf{P}$ to scattering and its ability to distinguish color-cast types by computing PSNR and SSIM between the corresponding representations of each degraded observation and its paired clean reference. 

\begin{figure}[htbp!]
    \centering
    \includegraphics[
    width=\linewidth
]{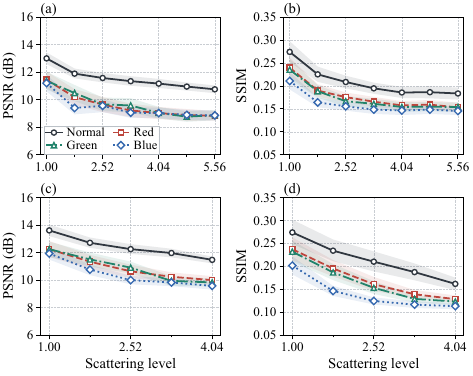}
    \caption{Similarity between degraded polarization
    representations and paired clean references on PURB.
    Panels (a) and (b) show PSNR and SSIM at an acquisition
    distance of 45~cm; panels (c) and (d) show the results
    at 60~cm. Curves represent normal, red-cast, green-cast,
    and blue-cast conditions.
    Shaded regions indicate pointwise 95\% confidence intervals
    obtained through scene-level bootstrap resampling.}
    \label{fig:full_polarization_vs_gt}
\end{figure}

\begin{figure}[htbp!]
    \centering
    \includegraphics[
    width=\linewidth
]{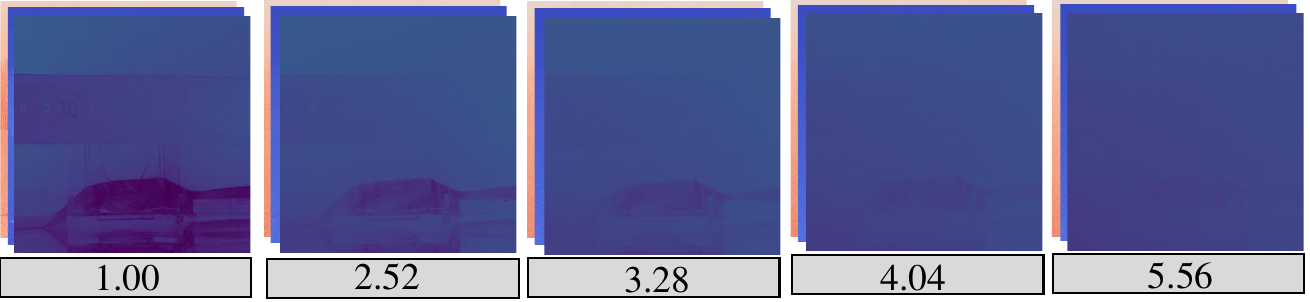}
    \caption{Variation in polarization characteristics with scattering severity}
    \label{fig:polarization_vs_scater}
\end{figure}

As shown in Fig.~\ref{fig:full_polarization_vs_gt}, the scene-averaged PSNR and SSIM generally decrease as scattering increases at both acquisition distances.
The qualitative comparisons in Fig.~\ref{fig:polarization_vs_scater} further support this trend, showing an overall decline in the similarity between the polarization representation and the reference images as scattering increases.
Together, these quantitative and qualitative results support the sensitivity of the polarization representation to scattering-related changes, although the quantitative metrics do not decrease strictly monotonically at intermediate scattering levels.

The separation between color-cast types is less pronounced.
The condition without a color cast yields higher mean
similarity to the clean reference, whereas the red-cast
and green-cast conditions follow closely spaced curves.
Under stronger scattering, the PSNR curves for the three
color-cast conditions become increasingly close. These observations suggest that
the scalar similarity measures provide limited
discrimination among specific color-cast types in this
analysis, motivating the joint use of polarization
information and RGB appearance cues for degradation
identification.
\subsection{Degradation Analysis on Mixpolar}
To characterize the empirical capability boundaries of
current open-source VLMs in underwater degradation perception
and the use of polarization information, we compare
PolarDU with three representative models on MixPolar:
Qwen3-VL-8B~\cite{bai2025qwen3},
InternVL3.5-8B~\cite{wang2025internvl3}, and
MiniCPM-V 4.5~\cite{yu2026minicpm}.
Each baseline is directly evaluated with RGB-only and RGB-Polar inputs under the same evaluation protocol and inference settings as PolarDU to assess the correctness of joint degradation diagnoses and whether adding polarization information improves attribute identification.

As shown in Table~\ref{tab:performance_comparison}, PolarDU
outperforms all compared VLMs in both joint accuracy
and macro-F1, reflecting more reliable joint identification
of degradation attributes and stronger class-balanced
recognition performance. Adding polarization inputs yields
inconsistent benefits: InternVL3.5-8B improves across the
reported metrics, whereas Qwen3-VL-8B shows mixed changes
and MiniCPM-V 4.5 performs worse. These results suggest
limitations in the use of polarization information by the
evaluated VLMs under the tested input settings, motivating
further exploration of targeted fine-tuning and cross-modal
alignment to improve their understanding of polarization
cues and degradation attributes.
\begin{figure}[htbp!]
\centering
\includegraphics[width=3.3in]{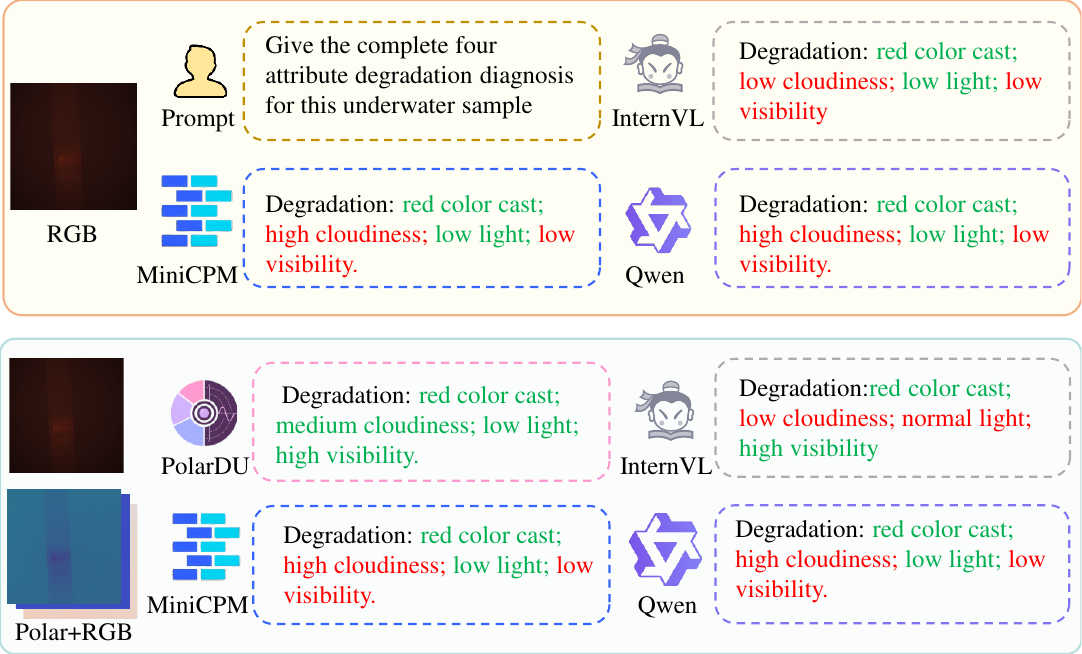}
\caption{Qualitative comparison of degradation identification
using RGB-only and RGB-Polar inputs. Green and red text
indicate correct and incorrect attribute predictions.}
\label{question}
\end{figure}

\begin{figure*}[htbp]
    \centering
    \includegraphics[width=\textwidth]{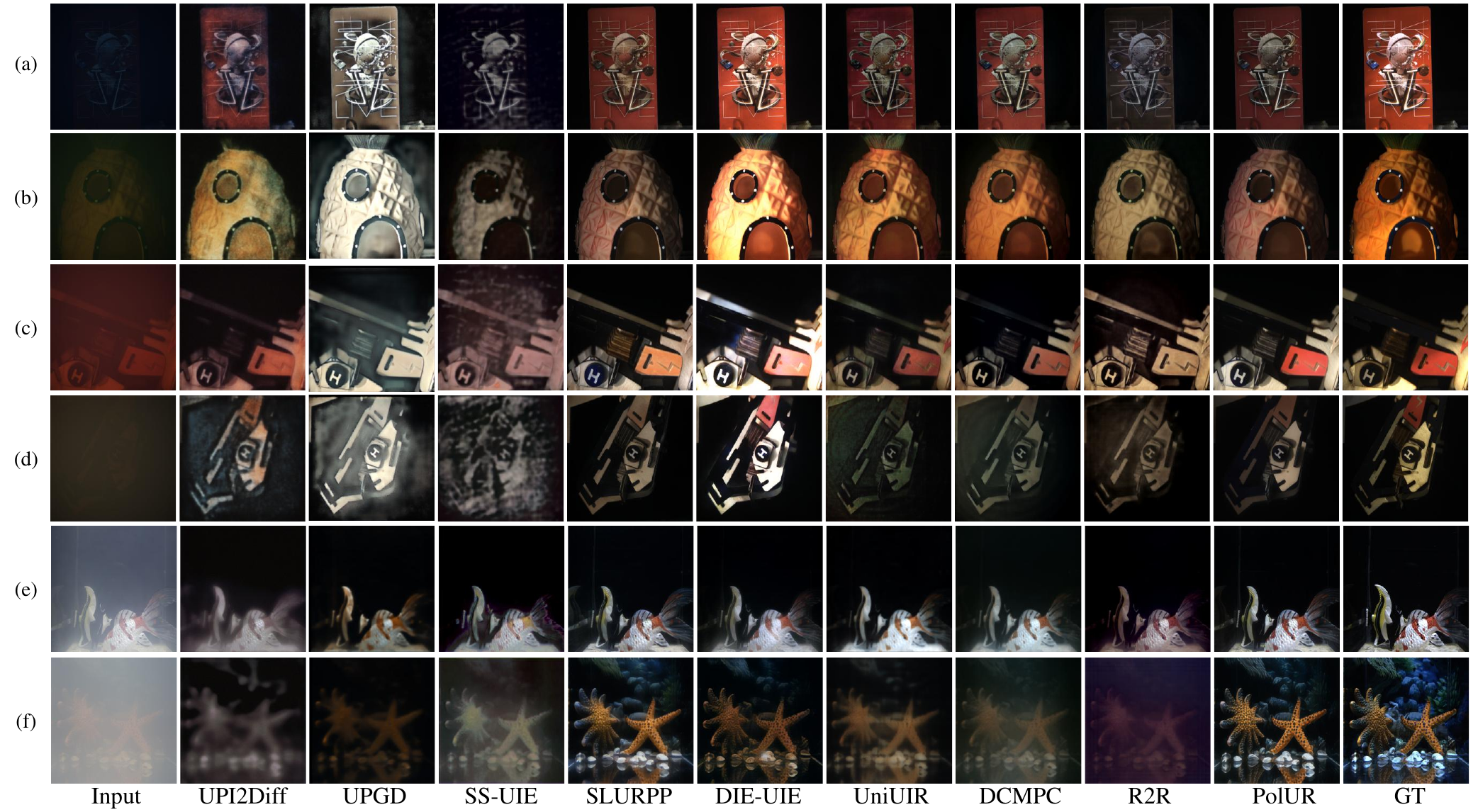}
    \caption{Visual comparisons on the MixPolar indoor subset and RGBP. Input and reference images (GT) are shown in the first and last columns, respectively.}
    \label{fig:comparison_indoor_rgbp}
\end{figure*}

To further illustrate the differences in joint degradation
identification, we qualitatively compare representative
RGB-only and RGB-Polar predictions, as shown in
Fig.~\ref{question}. All three compared VLMs correctly
identify the red color cast but misclassify cloudiness:
InternVL3.5-8B underestimates it, whereas Qwen3-VL-8B
and MiniCPM-V 4.5 overestimate it. Although adding
polarization inputs corrects the visibility prediction
of InternVL3.5-8B, it introduces an illumination error.
The other two models retain the same diagnoses under
both input settings and continue to underestimate
visibility. By comparison, PolarDU correctly identifies
all four attributes, providing a complete and accurate
degradation diagnosis for this example.
\begin{table}[t]
    \centering
    \caption{Performance Comparison of Different Methods}
    \label{tab:performance_comparison}
    \footnotesize
    \setlength{\tabcolsep}{3pt}
    \renewcommand{\arraystretch}{1.15}
    \begin{tabular}{@{}llrrr@{}}
        \toprule
        Model & Input & Accuracy (\%) & F1 (\%) & Score \\
        \midrule
        \multirow{2}{*}{Qwen3-VL-8B}
            & RGB-only  & 4.40 & 26.01 & 15.20 \\
            & RGB-Polar &1.70 & 26.96 & 14.33 \\
        \addlinespace[3pt]
        \multirow{2}{*}{InternVL3.5-8B}
            & RGB-only  & 1.31 & 32.62 & 16.97 \\
            & RGB-Polar & 7.33 & 37.69 & 22.51 \\
        \addlinespace[3pt]
        \multirow{2}{*}{MiniCPM-V 4.5}
            & RGB-only  & 6.56 & 30.28 & 18.42 \\
            & RGB-Polar & 6.02 & 24.45 & 15.23 \\
        \addlinespace[3pt]
        PolarDU
            & RGB-Polar
            & \textbf{76.23}
            & \textbf{90.07}
            & \textbf{83.15} \\
        \bottomrule
    \end{tabular}
\end{table}

\subsection{Restoration Performance in Controlled Scenes}

We compare PolUR with nine representative approaches, all evaluated using their official implementations and grouped into three categories: (1) polarization-based underwater restoration and enhancement methods, including UIRPIOM~\cite{li2024underwater}, UPI2Diff~\cite{UPI2Diff}, and UPGD~\cite{UPGD}, (2) underwater image restoration methods, including SS-UIE~\cite{SSUIE}, SLURPP~\cite{SLURPP}, and DIE2UIE~\cite{DIE2UIE}, and (3) all-in-one image restoration methods, including UniUIR~\cite{uniuir}, DCMPC~\cite{fan2026degradation}, and R2R~\cite{R2R}. We use their official implementations, pretrained checkpoints, and default configurations without further training or tuning.\par

\begin{table*}[htbp]
    \centering
    \caption{Quantitative comparison on the MixPolar indoor and RGBP. The best and second-best distinct values are marked in bold and underlined, respectively, tied best results are both marked in bold. $\uparrow$ and $\downarrow$ indicate that higher and lower values are better, respectively.}
    \label{tab:comparison_mixpolar_rgbp}
    \footnotesize
    \setlength{\tabcolsep}{4pt}
    \renewcommand{\arraystretch}{1.15}
    \begin{tabular*}{\textwidth}{@{\extracolsep{\fill}} l *{8}{c} @{}}
        \toprule
        \multirow{2}{*}{Method}
        & \multicolumn{4}{c}{MixPolar Indoor}
        & \multicolumn{4}{c}{RGBP} \\
        \cmidrule(lr){2-5}\cmidrule(lr){6-9}
        & PSNR $\uparrow$ & SSIM $\uparrow$ & LPIPS $\downarrow$ & DISTS $\downarrow$
        & PSNR $\uparrow$ & SSIM $\uparrow$ & LPIPS $\downarrow$ & DISTS $\downarrow$ \\
        \midrule
        UIRPIOM  & 6.6200 & 0.3030 & 0.7930 & 0.6570 & 9.3662 & 0.2467 & 0.6636 & 0.4265 \\
        UPI2Diff & 18.9040 & 0.4720 & 0.4236 & 0.3550 & 17.0288 & 0.3937 & 0.7035 & 0.4924 \\
        UPGD     & 14.2305 & 0.5081 & 0.3387 & 0.2879 & 15.8287 & 0.5195 & 0.4854 & 0.4589 \\
        SS-UIE   & 16.8250 & 0.4300 & 0.5101 & 0.3910 & 16.9519 & 0.1958 & 0.4805 & 0.3270 \\
        SLURPP   & \underline{20.1621} & \underline{0.7340} & \underline{0.1760} & \underline{0.1620}
                 & \underline{19.4897} & \underline{0.6037} & \underline{0.2214} & \underline{0.1800} \\
        DIE-UIE  & 17.3200 & \textbf{0.7350} & 0.1831 & \textbf{0.1610} & 18.6579 & 0.5225 & 0.3525 & 0.3235 \\
        UniUIR   & 19.1524 & 0.6290 & 0.2804 & 0.2390 & 18.1337 & 0.5898 & 0.4544 & 0.3583 \\
        DCMPC    & 19.7740 & 0.5620 & 0.2120 & 0.2100 & 18.3306 & 0.4982 & 0.4860 & 0.3890 \\
        R2R      & 18.5353 & 0.4120 & 0.2605 & 0.2470 & 16.2180 & 0.1822 & 0.4706 & 0.3870 \\
        \midrule
        \textbf{PolUR} & \textbf{20.8262} & \textbf{0.7350} & \textbf{0.1684} & 0.1680
                        & \textbf{19.7495} & \textbf{0.6181} & \textbf{0.2095} & \textbf{0.1609} \\
        \bottomrule
    \end{tabular*}
\end{table*}
As shown in Table~\ref{tab:comparison_mixpolar_rgbp}, PolUR
achieves the best or joint-best results on most metrics
across both datasets, demonstrating strong overall
restoration performance. Specifically, it leads in PSNR
and LPIPS while attaining the best or joint-best SSIM,
reflecting advantages in reconstruction fidelity,
structural consistency, and perceptual similarity to the
reference images. Although its DISTS score on the MixPolar
indoor subset is slightly higher than those of DIE-UIE
and SLURPP, the absolute gaps remain small (0.007 and
0.006, respectively).
\begin{table}[htbp!]
    \caption{Quantitative comparison on MixPolar Outdoor. The best and second-best results are marked in bold and underlined, respectively.}
    \label{tab:quantitative_comparison}
    \centering
    \footnotesize
    \setlength{\tabcolsep}{6pt}
    \renewcommand{\arraystretch}{1.15}
    \begin{tabular}{lccc}
        \toprule
        Method & UIQM $\uparrow$ & URanker $\uparrow$ & UCIQE $\uparrow$ \\
        \midrule
        UIRPIOM  & 0.6370          & -0.7039         & 0.3187 \\
        UPI2Diff & \textbf{2.9569} & -0.0010         & 0.3451 \\
        UPGD     & 1.8900          & 0.7613          & 0.2763 \\
        SS-UIE   & 2.2645          & 0.7897          & 0.3108 \\
        SLURPP   & 2.5210          & 0.3099          & 0.2473 \\
        DIE-UIE  & 2.7413          & 0.7794          & \underline{0.3472} \\
        UniUIR   & 2.7480          & \underline{1.0174} & 0.3370 \\
        DCMPC    & 2.0294          & 0.1379          & 0.3362 \\
        R2R      & 2.2145          & 0.1666          & 0.2912 \\
        \textbf{PolUR} & \underline{2.8934} & \textbf{1.2767} & \textbf{0.3529} \\
        \bottomrule
    \end{tabular}
\end{table}
We further conduct qualitative comparisons, as shown in Fig.~\ref{fig:comparison_indoor_rgbp}(b). Polarization-based optimization methods are omitted from the figure due to their poor restoration quality. Other polarization-based restoration methods recover some structural details but still exhibit noticeable color casts and blurring. Among the image restoration methods, SS-UIE exhibits incomplete detail recovery and grainy artifacts. SLURPP preserves clear overall structures, although the colors in some regions still differ from the reference images. DIE-UIE produces local overexposure in some examples, reducing detail visibility. In examples with pronounced haze, all-in-one restoration methods, including UniUIR, DCMPC, and R2R, still exhibit blurred details, residual haze, or color casts. By comparison, PolUR preserves clearer textures and object contours, achieving a good balance between detail recovery and color cast suppression.
\subsection{Generalization to Real-World Underwater Scenes}

To assess PolUR’s generalization ability, we directly evaluate it on the MixPolar outdoor subset, which was excluded from both its degradation identification and image restoration training. Given the distribution differences between indoor laboratory environments and real-world outdoor scenes, underwater restoration baselines use their official pretrained checkpoints for a fair assessment of outdoor generalization, while baselines requiring adaptation to underwater restoration are trained on the MixPolar indoor subset. All methods are evaluated directly on the outdoor subset without further training or fine-tuning.

As shown in Table~\ref{tab:quantitative_comparison}, PolUR
achieves the highest URanker and UCIQE scores, performing
well under both learned quality ranking and statistical
assessment of color and contrast. Although it ranks second
in UIQM, the absolute gap to UPI2Diff is only 0.0635.
These complementary results support PolUR's competitive
restoration performance in the tested real-water conditions
without additional training.

\begin{figure*}[htbp!]
    \centering
    \includegraphics[width=\textwidth]{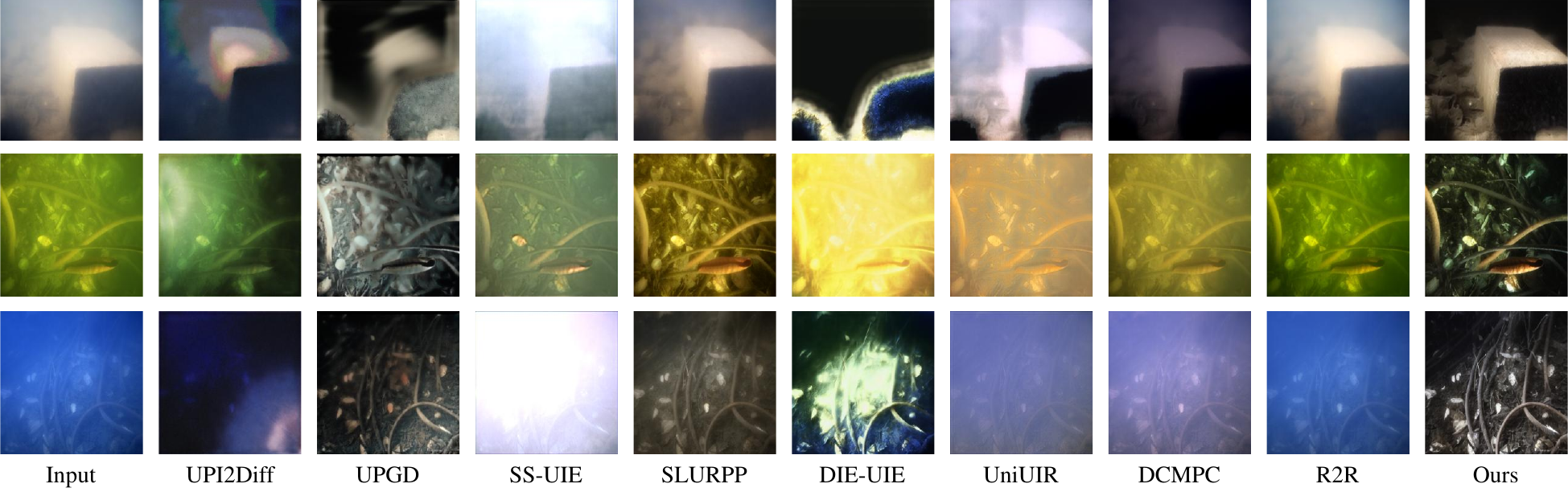}
    \caption{Visual comparisons on real-world outdoor scenes from MixPolar under direct evaluation without target-domain retraining or fine-tuning. The first and last columns show the input images and PolUR results, respectively.}
    \label{fig:comparison_outdoor}
\end{figure*}

The visual results in Fig.~\ref{fig:comparison_outdoor} complement these scores. UPI2Diff retains residual
color casts and leaves some details in dark regions
indistinct, while UPGD reduces the casts but produces
relatively smooth surface details in the first example.
Among the image restoration methods, SS-UIE produces
excessively bright regions, SLURPP retains yellowish or
grayish appearances, and DIE-UIE exhibits yellow
amplification and local overexposure. All-in-one restoration
methods, including UniUIR, DCMPC, and R2R, retain residual
haze or much of the original color cast, limiting scene
visibility. In comparison, PolUR reduces the dominant
green and blue casts and makes the fish, vegetation, and
surrounding surfaces more distinguishable, although
details in some shadow regions remain limited.

\subsection{Ablation Studies}
We conduct ablation studies on degradation identification
and image restoration to assess the contributions of
the main design choices in PolUR.
\subsubsection{Degradation Identification}

To assess the contributions of the input modalities and
the polarization alignment stage, we conduct ablation
studies. We compare RGB-only and polarization-only variants
with the full model. The variant without stage 1 skips
polarization alignment and proceeds directly to joint
fine-tuning.

As shown in Table~\ref{tab:ablation_identification},
the full model achieves the highest  joint accuracy
and macro-F1, outperforming both single-modality variants
and supporting the benefit of jointly using RGB and
polarization information. Although the RGB-only variant
performs better than the polarization-only variant,
its lower performance than the full model suggests that
polarization information provides complementary cues
for degradation identification. Removing stage 1 also
reduces both metrics despite retaining joint fine-tuning,
supporting the contribution of polarization alignment
to complete attribute diagnosis and class-balanced
recognition.

\begin{table}[t]
    \centering
    \caption{Ablation of input modalities and the training
    strategy for degradation identification on MixPolar.
    Acc. denotes joint accuracy, and Score is the
    arithmetic mean of Acc. and macro-F1.
    Best and second-best results are highlighted in bold
    and underlined.}
    \label{tab:ablation_identification}
    \footnotesize
    \setlength{\tabcolsep}{4pt}
    \renewcommand{\arraystretch}{1.15}

    \begin{adjustbox}{max width=\linewidth}
        \begin{tabular}{@{}lccc@{}}
            \toprule
            Variant
            & Acc. (\%) $\uparrow$
            & Macro-F1 (\%) $\uparrow$
            & Score $\uparrow$ \\
            \midrule
            RGB only
            & 70.14 & 87.46 & 78.80 \\
            Polarization only
            & 56.02 & 79.59 & 67.80 \\
            w/o stage 1
            & \underline{72.07}
            & \underline{88.23}
            & \underline{80.15} \\
            \midrule
            Full model
            & \textbf{76.23}
            & \textbf{90.07}
            & \textbf{83.15} \\
            \bottomrule
        \end{tabular}
    \end{adjustbox}
\end{table}

\subsubsection{Image Restoration}

To assess the contributions of the restoration components,
we construct three ablation variants: removing RPA,
omitting degradation guidance from PolarDU
(w/o recognition), and replacing the adapter with
direct pooling to match the required feature dimensions
(w/o adapter). All variants are evaluated on the
MixPolar indoor subset.

As shown in Table~\ref{tab:ablation_restoration},
omitting the hidden states from PolarDU or removing RPA
reduces both PSNR and SSIM, supporting the use of degradation
guidance and the alignment and fusion of VAE-encoded
polarization and RGB features for restoration.
The adapter learns to compress the hidden states into
compact conditioning features and achieves higher PSNR
than direct pooling, although its lower SSIM reveals
a trade-off between reconstruction fidelity and
structural similarity.

\begin{table}[t]
    \centering
    \caption{Ablation of image restoration components
    on the MixPolar indoor subset.
    Best results are shown in bold, and second-best
    results are underlined.}
    \label{tab:ablation_restoration}
    \footnotesize
    \setlength{\tabcolsep}{6pt}
    \renewcommand{\arraystretch}{1.15}

    \begin{adjustbox}{max width=\linewidth}
        \begin{tabular}{@{}lcc@{}}
            \toprule
            Variant
            & PSNR (dB) $\uparrow$
            & SSIM $\uparrow$ \\
            \midrule
            w/o recognition
            & 20.3990 & 0.728 \\
            w/o RPA
            & 20.4408 & 0.725 \\
            w/o adapter
            & \underline{20.5936}
            & \textbf{0.746} \\
            \midrule
            Full model
            & \textbf{20.8262}
            & \underline{0.735} \\
            \bottomrule
        \end{tabular}
    \end{adjustbox}
\end{table}
\subsection{Downstream Segmentation Evaluation}
To assess whether PolUR enhancement improves downstream binary semantic segmentation, we train two LR-ASPP~\cite{howard2019searching} models on raw underwater images and their enhanced counterparts, respectively. Both models use identical initialization, data splits, reference masks, and training settings.
As shown in Table~\ref{tab:downstream_segmentation},
LR-ASPP achieves higher IoU and Dice and lower MAE than its
counterpart using raw images, reflecting better agreement
with the reference masks and reduced pixel-wise prediction
error.

\begin{table}[t]
    \centering
    \caption{Downstream segmentation performance of
    SegFormer trained and evaluated on raw and
    PolUR-enhanced images. Best results are shown
    in bold.}
    \label{tab:downstream_segmentation}
    \footnotesize
    \setlength{\tabcolsep}{5pt}
    \renewcommand{\arraystretch}{1.15}

    \begin{adjustbox}{max width=\linewidth}
        \begin{tabular}{@{}lccc@{}}
            \toprule
            Images
            & IoU (\%) $\uparrow$
            & Dice (\%) $\uparrow$
            & MAE $\downarrow$ \\
            \midrule
            Raw
            & 83.10 & 90.16 & 0.006005 \\
            Enhanced
            & \textbf{83.22}
            & \textbf{90.23}
            & \textbf{0.005870} \\
            \bottomrule
        \end{tabular}
    \end{adjustbox}
\end{table}
\begin{figure}[htbp!]
    \centering
    \includegraphics[width=3.3in]{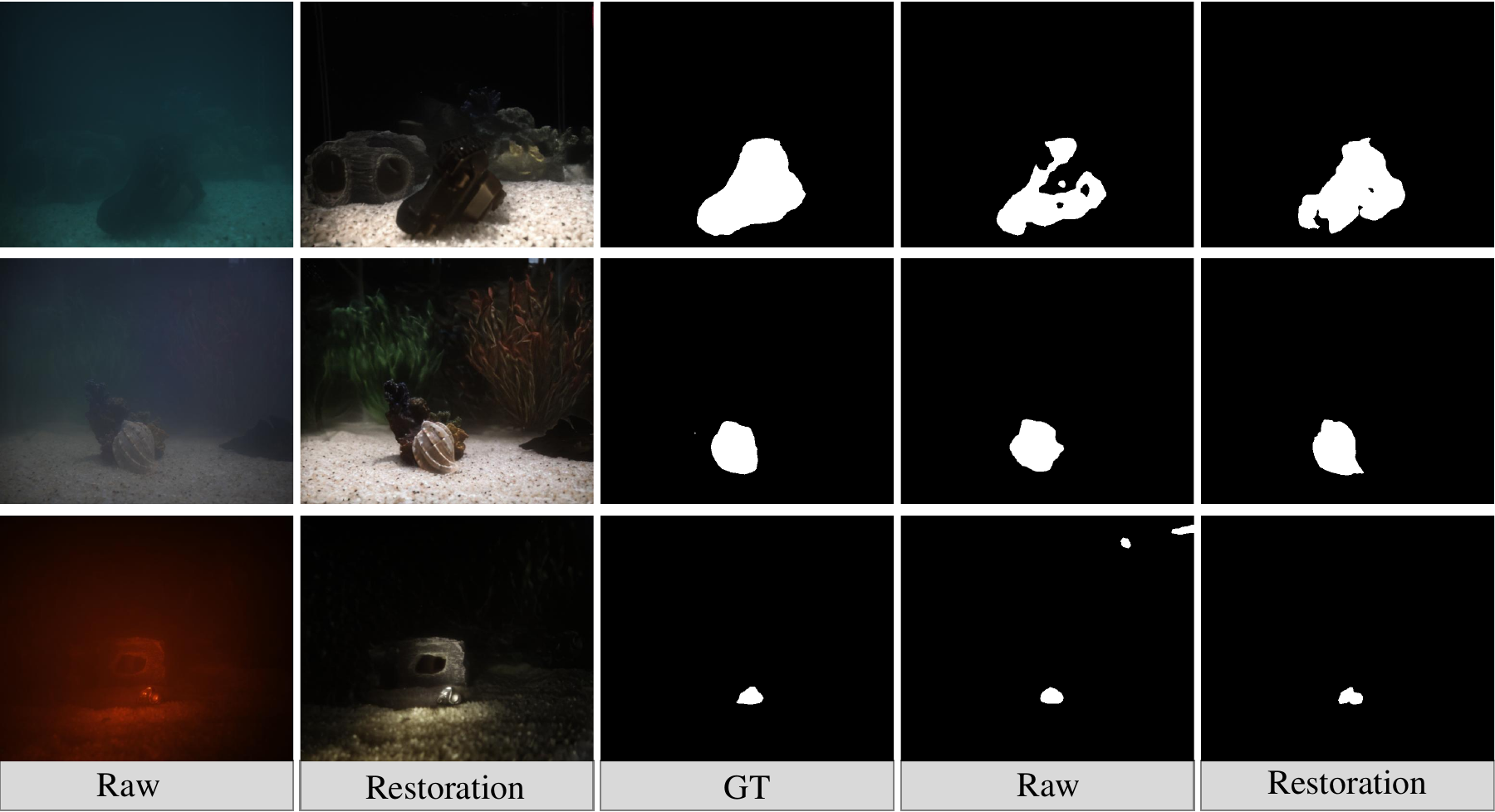}
    \caption{Visual comparison of downstream segmentation
    results. From left to right: raw images, PolUR-restored
    images, ground-truth masks (GT), and predictions from
    SegFormer trained and evaluated on raw and restored
    images, respectively.}
    \label{fig:downstream_segmentation}
\end{figure}
We further analyze representative examples
in Fig.~\ref{fig:downstream_segmentation}. In the first
example, the restored-image setting produces a more
complete foreground mask with fewer internal gaps.
In the second example, the predicted contour more closely
matches the reference mask, while in the third, isolated
false-positive regions in the upper-right background are
removed. These examples complement the quantitative
results by illustrating improved mask completeness and
reduced background false positives, although some
boundary discrepancies remain.
\section{Conclusion}
In this paper, we present PolUR, a unified framework for underwater degradation understanding and image restoration. Guided by textual instructions, PolUR uses complementary RGB and polarization information to understand mixed degradations and restore degraded images. We further developed a progressive training strategy and an RPA module to reduce cross-modal representation discrepancies during degradation perception and restoration. Meanwhile, we also constructed MixPolar, a multi-task benchmark supporting degradation understanding, image restoration, and downstream segmentation evaluation. Extensive experiments demonstrate that PolUR achieves competitive performance in both degradation understanding and image restoration.

\bibliographystyle{IEEEtranS}

\bibliography{IEEEabrv,ref}

\end{document}